\pdfoutput=1

\documentclass[11pt]{article}

\usepackage[]{emnlp2021}

\usepackage{times}
\usepackage{latexsym}

\usepackage{stfloats}

\usepackage{nert}

\usepackage{microtype}

\title{CoreLM: Coreference-aware Language Model Fine-Tuning}

\author{Nikolaos Stylianou \and Ioannis Vlahavas\\
  Aristotle University of Thessaloniki \\
  School of Informatics \\
  Thessaloniki, Greece \\
  \texttt{\{nstylia,vlahavas\}@csd.auth.gr}

  }

\begin{document}
\maketitle
\begin{abstract}
Language Models are the underpin of all modern Natural Language Processing (NLP) tasks. The introduction of the Transformers architecture has contributed significantly into making Language Modeling very effective across many NLP task, leading to significant advancements in the field. However, Transformers come with a big computational cost, which grows quadratically with respect to the input length. This presents a challenge as to understand long texts requires a lot of context. In this paper, we propose a Fine-Tuning framework, named CoreLM, that extends the architecture of current Pretrained Language Models so that they incorporate explicit entity information. By introducing entity representations, we make available information outside the contextual space of the model, which results in a better Language Model for a fraction of the computational cost. We implement our approach using GPT2 and compare the fine-tuned model to the original. Our proposed model achieves a lower Perplexity in GUMBY and LAMBDADA datasets when compared to GPT2 and a fine-tuned version of GPT2 without any changes. We also compare the models' performance in terms of Accuracy in LAMBADA and Children's Book Test, with and without the use of model-created coreference annotations.
\end{abstract}

\section{Introduction}
Language Models (LMs) have seen significant improvements in performance due to the Transformers architecture \cite{vaswani2017transformers}. The resulting Pretrained Language Models (PLMs) such as BERT \cite{devlin-etal-2019-bert}, GPT2 \cite{radford2019language} and XLNet \cite{yang2019xlnet} have contributed to significant advancements in many Natural Language Processing (NLP) tasks. PLMs, regardless of their training objective and methodology, aim to learn contextualized text representations. As such, the available context during training has a key role in the models performance.

The quadratic computation complexity in the attention mechanism of the Transformer architecture, in terms of input sequence length, has been a limiting factor to the amount of contextual information the models can have at each step. To make this architecture more efficient, a plethora of approaches have been introduced \citep[][\textit{inter alia}]{kitaev2020reformer,Beltagy2020Longformer,child2019generating}, aiming to lower the computational complexity without sacrificing performance. \citet{tay2020efficient} summarizes these approaches and categorizes them based on type of attention mechanism in their survey. Yet, even with linear complexity, physical resources will always limit the amount of contextual information that can be handled simultaneously. 

In terms of language, entities represent a natural way to tie words together through large pieces of discource. In NLP, Coreference Resolution is the task that aims to identify and group these entity mentions together, when they refer to the same real world entity \cite{stylianou2021coref}. Therefore, Coreference Resolution presents a natural way to link the context that we are currently handling with distant information, far outside the capabilities model architectures and current hardware resources. 
However, while this information is present in the text, it is usually not annotated and when annotated are very sparse and in small quantities making it extremely difficult to train large LMs \cite{kunz-hardmeier-2019-entity}. 

In this paper we present a framework to effectively use coreference annotations to further fine-tune large PLMs, increasing their performance far more than just by fine-tuning on the same data. By using large PLMs we take advantage of existing resources that are expensive to reproduce and come with a big environmental cost \cite{strubell-etal-2019-energy}. What is more, fine-tuning takes advantage of the massive amount of data that essentially initialize the model, making it possible to introduce new capabilities to models with small amounts of annotated data.

In our approach, we use GPT2 as our base model and extend its architecture with the addition of a new Entity-Gating layer that handles entity annotations along with a gating mechanism that handles information flow between the base model and the Entity-Gating layer. As such, our approach uses entity representations when they are available through both training and inference, without imposing any constraints to the model's functionality. 

For our experiments, we compare the performance of GPT2, post and pre fine-tuning, with and without our changes in a series of relative tasks. Furthermore, since most coreference annotated datasets are either very small or hard to acquire, we use GUMBY \cite{gessler2020gumby} as our fine-tuning dataset which is a model annotated corpus. In addition, by using noisy annotations we aim to show the resilience of our approach to noise and the universality of our framework. Our results highlight the effects of our framework in language modeling, modeling long-range dependencies, and in specific word types where our fine-tuned model achieves better performance than GPT2.

\section{Background}
\label{sec:background}
This section provides a concise overview of the Transformers architecture \cite{vaswani2017transformers}, which is the foundation of our approach, followed by a brief explanation of autoregressive language modeling used by our base model, GPT2. 
\subsection{Transformers}
\label{sec:vanilla_transformers}
The Transformers architecture is based on stacked Transformer blocks, which take as input a $k \times d$ input vector and return a same size vector after applying a sequence of operations, where $k$ and $d$ denote the context window size and hidden size respectively. Each block is consisted of a multi-head masked self attention layer and a two layer position-wise feed-forward network, each rapped with a layer normalization (\texttt{LayerNorm}) layer \cite{ba2016layer} and a residual connection \cite{he2016deep}. Formally, given an input $X$ the encoder and decoder architecture is described as: 

\noindent \texttt{Encoder:}
\begin{equation}
\label{eq:vanilla_transformers_encoder}
\begin{aligned}
    Y = \texttt{LayerNorm}(\texttt{Self-Attention}(X)) + X \\
    Z = \texttt{LayerNorm}(\texttt{PositionFFN}(Y)) + Y 
\end{aligned}
\end{equation}

\noindent \texttt{Decoder:}
\begin{equation}
\label{eq:vanilla_transformers_decoder}
\begin{aligned}
   T = \texttt{LayerNorm}(\texttt{Self-Attention}(T)) + T \\
   P = \texttt{LayerNorm}(\texttt{Self-Attention}(T,Z)) + T \\
   H = \texttt{LayerNorm}(\texttt{PositionFFN}(P)) + P 
\end{aligned}
\end{equation}

However, the decoder can also be used independently by eliminating the second Self-Attention layer. 
\paragraph{Self-Attention:}
The self-attention mechanism takes as input a vector $X$ and projects it into $Q$, $K$, $V$ representations for the Query, Key, Value attention scheme. Using the projected vector, this mechanism is formalized as: 
\begin{equation}
\label{eq:vanilla_self_attention}
    \texttt{Self-Attention} = \texttt{softmax}(\frac{Q K}{\sqrt{d}})V
\end{equation}
where d is the size of the $Q$, $K$, $V$ vectors. Usually the self-attention is multi-headed, in which multiple attentions are calculated in parallel, with the outputs of the multi-headed attentions being concatenated.
\paragraph{PositionFFN:}
Given an input vector $X$, this layer applies two position-wise linear transformations with a ReLU activation in between. The PositionFFN layer is formalized as: 
\begin{equation}
\label{eq:vanilla_positionffn}
    \texttt{PositionFFN}=max(0, X W_{1} + b_{1})W_{2} + b_{2}
\end{equation}
with $W_{1},b_{1}$ and $W_{2},b_{2}$ being the trainable weight and bias parameters of each layer respectively.

\subsection{Autoregressive Language Modeling}
\label{sec:autoregressive_lm}
Autoregressive language models estimate the distribution over a sequence of word tokens by factorizing their joint probabilities as the product of conditional probabilities \cite{bengio2003neural}. For a context vector of tokens $U = (u_{1}, \dots, u_{k})$, this is formally described as:   

\begin{equation}
\label{eq:lm_prob}
    p(U) = \prod_{i=1}^{k}p(u_{k} | u_{1}, \dots , u_{k-1})
\end{equation}

where, $k$ is the context window size. 

GPT2 is an autoregressive LM, based on the previously described Transformer's decoder architecture.  In comparison to the previously described architecture, it uses masked multi-headed attention to prevent leftward information flow in the decoder. For further information about the model's architecture we refer readers to \citet{radford2019language}.

\section{Coreference-aware Language Modeling}
Language models are limited to the amount of information they can process based on the available context window $k$ (\cref{eq:lm_prob}). However, increasing $k$ will lead to an exponential increase in computation complexity. As such, we implicitly increase the available information, without increase the context window, using coreference annotations. 

We do this by introducing entity-representations (\cref{sec:entity_representations}), in the form of vectors, that are utilized by the model to infuse the respective entity information to all the tokens in the sequence that are part of entities. The entity-representations are created from the whole discourse available, hence holding contextual information that are very distant to the context window. As such, we make no alterations to the language modeling objective. These vector representations are introduced to the model via an Entity-Gating layer (\cref{sec:entity_gating}) that is added to model architecture. 

\subsection{Architecture}
\label{sec:architecture}
Our base model, GPT2, is comprised of $N$ stacked Transformer decoder blocks, where each is consisted of a multi-headed attention layer and a position-wise feedforward layer with residual connections and layer normalizations. We extend its architecture by adding an Entity-Gating layer after the Transformer decoders (\cref{eq:gpt_input,eq:gpt2_base,eq:entity_gating_arc,eq:gpt2_output}).
\begin{equation}
\label{eq:gpt_input}
     h_{0} = UW_{e} + W_{p}
\end{equation}
\begin{equation}
\label{eq:gpt2_base}
     h_{l} = \texttt{transformer\_decoder}(h_{l-1}) \forall_{i} \in [1,n]  \\
\end{equation}
\begin{equation}
\label{eq:entity_gating_arc}
     h_{e} = \texttt{entity\_gating}(h_{n},E) \\ 
\end{equation}
\begin{equation}
\label{eq:gpt2_output}
     p(U) = \texttt{softmax}(h_{e}W_{e}^{T})
\end{equation}

\begin{figure}[!ht]
    \centering
    \includegraphics[scale=0.5]{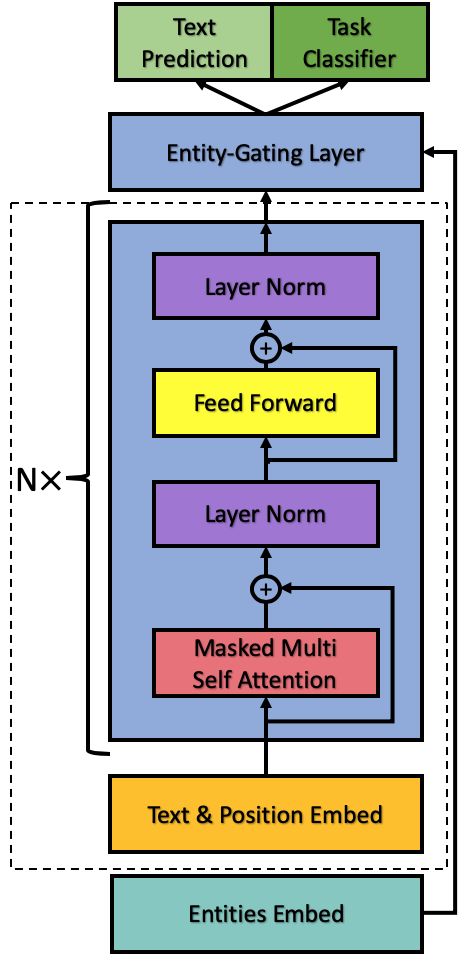}
    \caption{CoreLM Architecture as an extention of GPT2 model (area within dashed border).}
    \label{fig:corelm}
\end{figure}

Here, $n$ is the number of layers, $W_{e}$ is the token embedding matrix, $W_p$ is the position embedding matrix and $E$ is the context vector of entity representations. \Cref{fig:corelm} illustrates a high level view of the model of our proposed architecture.

\subsection{Entity representations}
\label{sec:entity_representations}
Each entity is represented by a learned vector $E_{i} \in \mathbb{R}^{1 \times d_{embd}}$, where $d_{embd}$ is the embedding dimension of the model ($W_{e}$). These entity vectors are stored in a persistent set of entities $\mathcal{E}$ so that they can be utilized through out the whole discourse scope. We use $E_{0}$ as a static entity representation for tokens that are not part of an entity.

The entity representations are initialized as a vector of ones. This design choice is based on the architecture of the Entity-Gating layer (\cref{sec:entity_gating}) and the learning process. Specifically, as each token is accompanied with an entity representation. Initializing the entity representations this way introduces less noise during the first occurrence of an entity mention. It also provides a dynamic way of using the same architecture, even when no entity are available. 

\subsection{Entity-Gating}
\label{sec:entity_gating}

\begin{figure}[!ht]
    \centering
    \includegraphics[scale=0.5]{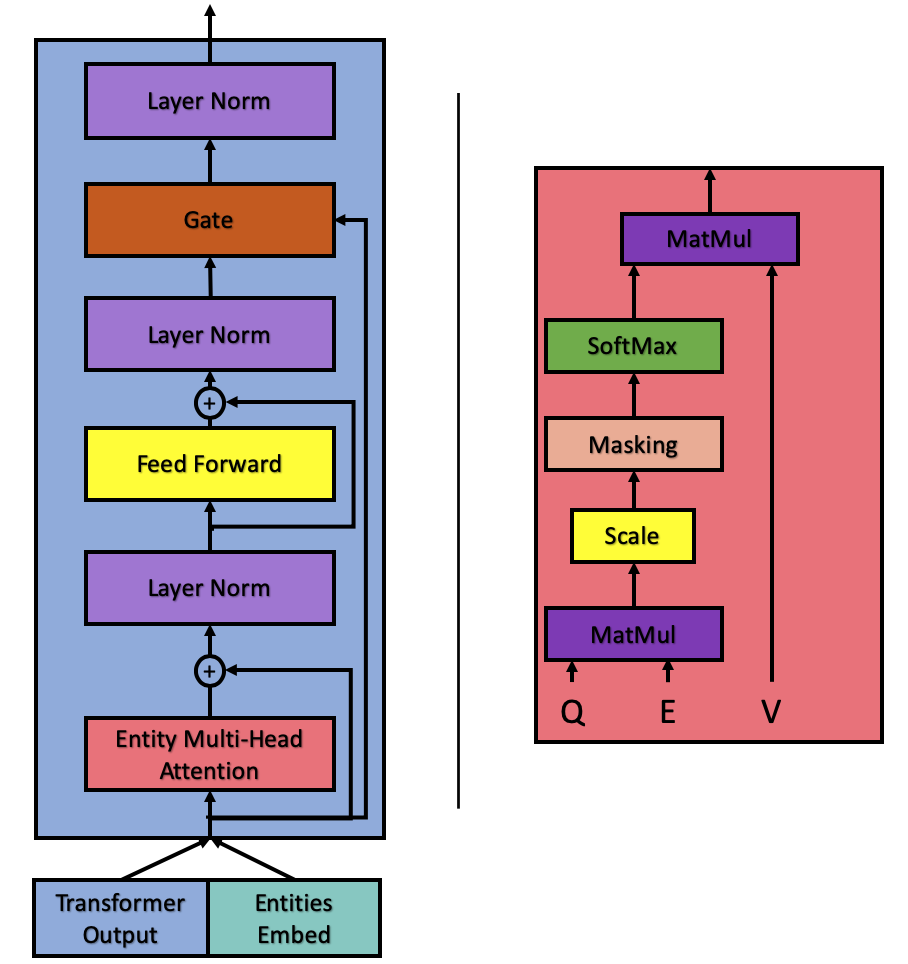}
    \caption{\textbf{(left)} Entity-Gating layer architecture. \\\textbf{(right)} Entity-Attention mechanism. In the illustration, we assume a single attention head for simplicity.}
    \label{fig:entity_gating}
\end{figure}

Our proposed Entity-Gating layer (\Cref{fig:entity_gating}) follows the same design principles of the GPT2 Transformer decoder blocks, using a Multi-Headed Attention layer, Layer Normalization layers and residual connections. However, we replace the Masked Multi-Head Self Attention layer with an Entity-Attention layer and use a learnable gating mechanism to control flow of information. Formally, the Entity-Gating layer is described as:

\begin{equation}
\label{eq:entity_gating_attention}
    EG_{A} = \texttt{LayerNorm}(\texttt{EntityAttention}(h_{l})) + h_{l} \\
\end{equation}
\begin{equation}
\label{eq:transformer_ffn}
    EG_{B} = \texttt{LayerNorm}(\texttt{PositionFFN}(EG_{A})) + EG_{A}
\end{equation}
\begin{equation}
\label{eq:entity_gating_gate}    
    h_{e} = \texttt{LayerNorm}(\texttt{Gate}(EG_{B},h_{l}))
\end{equation}
The $\texttt{LayerNorm}$ and $\texttt{PositionFFN}$ are used as described in Section \ref{sec:vanilla_transformers}, from the original architecture.

Our \texttt{EntityAttention} layer uses the Query (Q), Entity (E), Value (V) scheme described in \cref{eq:entity_attention} \citep{stylianou-vlahavas-2020-e}. In comparison to their variation, we use a multi-headed approach so that we limit the effect of entity representation to the tokens closer to the entity mention. As such, the attention mechanism is defined as: 

\begin{equation}
\label{eq:entity_attention}
    EntityAttention = \texttt{Softmax}(\frac{Q E}{\sqrt{d_{k}}})V    
\end{equation}

where $d_{k}$ is the dimension of the queries and entities in each head. Finally, the $\texttt{Gate}$ layer combines the layer input $h_{l}$ with $EG_{B}$ using the following gating mechanism before applying a layer normalization to the output: 
\begin{equation}
    \mathrm{g_{e}} = \delta (\sigma (V h_{l}))
\end{equation}
\begin{equation}
    \mathrm{z_{e}} = (1-\mathrm{g_{e}})\odot EG_{B} + \mathrm{g_{e}} \odot h_{l}
\end{equation}

in which $\sigma$ is a sigmoid function, $V$ is a parameter vector, $\odot$ is the Hadamard product and $\delta$ is a gate flow rate. We use a gate flow rate to ensure that the entity information is considered by our final model. By definition, $\delta \xrightarrow[]{}0$ results in no pass-through of information from the gate, $\delta \xrightarrow[]{}1$ results in completely dependent pass-through from the learned vector and $\delta \xrightarrow[]{}0.5$ enforces at least a fifty-fifty split of information. 

\section{Experiments}
We Fine-Tune our entity-aware LM with the original training objective of maximizing the log-probability of $U$ (\cref{eq:lm_prob}). Language Models are evaluated in terms of Perplexity (PPL) which is the exponentiated average negative log probability per word prediction, as we are using a word-level base model. As such, PPL is a direct reflection of the model's loss. For Fine-Tuning we use GUMBY, a model annotated coreference corpus, containing 4960 documents. 

The entity-aware fine-tuned LM is evaluated on GUMBY, using the created entity representations during Fine-Tuning and in a Zero-Shot setting, without any further training, on LAMBADA, WikiText2, WikiText103 in terms of PPL and on LAMBADA and CBT in terms of Accuracy. We also evaluate the effect of newly introduced entity annotations from a separate Coreference Resolution model on the Zero-Shot evaluated corpora, to investigate their effect in the model's performance.
Detail information about the model parameters and experimental setup are provided in \Cref{sec:experimental_setup}. \Cref{sec:coreference_annotations} enlists our methodology to annotate the LAMBADA and CBT corpora with coreference clusters, used in Zero-Shot evaluation only. 

\subsection{Fine-Tuning}
In order to fine-tune the model, using the annotated entity information in the GUMBY corpus \cite{gessler2020gumby}, we re-formatted it by introducing a second input stream along the raw text, which assigns a unique entity identifier in the corresponding token in the text \citep{stylianou-vlahavas-2020-e}. 

We similarly use $\emptyset$ to identify tokens that are not part of an entity. However, in order to utilize the encapsulated entity information present, we create multiple instances of the source files each annotated with a single layer of entity annotations. This comes in comparison with their approach in which only the outer entities are considered. We treat all entities as encapsulated entities, so that the second instance of the source file is only annotated with the entities that were identified within the span of text of first entity layer (example in \Cref{sec:experimental_setup} - \Cref{tab:gumby_example}). As a result, our final data was consisted of 13070 documents, which is approximately triple the original size. As the original corpus does not provide with predefined split, we held out 10\% of the data for evaluation while maintaining the balance between source types. 

The fine-tuned model is compared against the base model (GPT2) and a fine-tuned version of the base model without any entity information, on the held out data of GUMBY. 
As shown in \Cref{tab:lm_results} our approach significantly improves the model's performance after fine-tuning compared to the original model. 

\subsection{Language Modeling}
\label{sec:experiments_lm}
For the Language Modeling evaluation we use WikiText2, WikiText103 \cite{merity2016pointer} and LAMBADA \cite{paperno-etal-2016-lambada} in a Zero-Shot evaluation setting. We compare the previously fine-tuned entity-aware LM with the base-model, with and without any Fine-Tuning on GUMBY. The results are showcased in \Cref{tab:lm_results}. 

\begin{table*}[!t]
\caption{Language Modeling performance evaluation on fine-tuning and zero-shot setting. The fine-tuned models are trained only on GUMBY and evaluated on LAMBADA, WikiText2 and WikiText103 test sets.}
\centering
\begin{tabular}{l|c||ccc}
\toprule
 & \multicolumn{1}{c||}{\begin{tabular}[c]{@{}c@{}}GUMBY\\ (PPL)\end{tabular}} & \multicolumn{1}{c}{\begin{tabular}[c]{@{}c@{}}LAMBADA\\ (PPL)\end{tabular}} & \multicolumn{1}{c}{\begin{tabular}[c]{@{}c@{}}WikiText2\\ (PPL)\end{tabular}} & 
 \multicolumn{1}{c}{\begin{tabular}[c]{@{}c@{}}WikiText103\\ (PPL)\end{tabular}} \\ \midrule
\begin{tabular}[c]{@{}l@{}}GPT2\\Zero-Shot\end{tabular}& 66.19 & 26.49 & \textbf{30.63} & 35.81\\  \midrule
\begin{tabular}[c]{@{}l@{}}GPT2\\+Fine-Tuning\end{tabular} & 52.60 & 28.93  & 32.36 & 35.53 \\  \midrule 
\begin{tabular}[c]{@{}l@{}}GPT2\\+CoreLM\\+Fine-Tuning\end{tabular} & \textbf{46.97} & \textbf{26.21} & 31.80 & \textbf{29.51}\\
\bottomrule
\end{tabular}
\label{tab:lm_results}
\end{table*}

We notice that fine-tuning GPT2 with GUMBY does not generalize well in other domains, leading in a significant drop in performance in both LAMBADA and WikiText2. In WikiText103, in which the unknown words are proper names, the fine-tuned version performs better. With CoreLM, our model avoids catastrophic forgetting and presents similar performance gains in all corpora, resulting in a slight improvement in LAMBADA and WikiText103 and a slight impairment in WikiText2 compared to the base model.

\subsection{LAMBADA}
\label{sec:lambada}
The LAMBADA corpus is used to evaluate our approach, in a Zero-shot setting. LAMBADA is designed  to test the model's ability to use long range dependencies in text. Long range dependency, in this context, is consider to be a context window of 50 tokens, which the architecture can handle due to the 1024 token window context. During evaluation, we use use model acquired coreference annotation to the corpus, using a pretrained model Coreference Resolution model (\Cref{sec:coreference_annotations}). Hence, we also evaluate the effect of coreferent information. 

\paragraph{Without Coreference:}
Comparing the model's performance to GPT2, our approach achieves increased Accuracy in correctly predicting the last word, with scores 46.67\% for GPT2 and 48.11\% for CoreLM (statistically significant with paired t-test $p << 0.01$ - \Cref{tab:acc_results}). This increase in performance is slightly bigger if we compare it to the fine-tuned GPT2 model on GUMBY.

\paragraph{With Coreference:}
Coreference annotations offer a slight increase in performance, with 0.28\% Accuracy increase compared to the CoreLM version without coreference annotations. While this increase is not a statistically significant contribution, it is an expected behavior given that the entity representations were initialized during the zero-shot evalutation (\Cref{sec:coreference_annotations}). Furthermore, the context of each of LAMBADA entries is well inside the capabilities of the GPT2 model architecture and consequently the coreference annotations did not provide any information that were not already accessible by the model.

\subsection{Children's Book Test}
\label{sec:cbt}

\begin{table*}[!ht]
\caption{Zero-Shot evaluation on LAMBADA and Children's Book Test (CBT) with and without coreference annotations.}
\centering
\begin{tabular}{lccccc}
\toprule
 & \multicolumn{1}{c}{\begin{tabular}[c]{@{}c@{}}LAMBADA\\ (Acc)\end{tabular}} & \multicolumn{1}{c}{\begin{tabular}[c]{@{}c@{}}CBT-CN\\ (Acc)\end{tabular}} & \multicolumn{1}{c}{\begin{tabular}[c]{@{}c@{}}CBT-NE\\ (Acc)\end{tabular}} &
 \multicolumn{1}{c}{\begin{tabular}[c]{@{}c@{}}CBT-V\\ (Acc)\end{tabular}} &
 \multicolumn{1}{c}{\begin{tabular}[c]{@{}c@{}}CBT-P\\ (Acc)\end{tabular}} \\\midrule
\begin{tabular}[c]{@{}l@{}}GPT2\\Zero-Shot\end{tabular}& 46.67 & \textbf{84.48} & {64.52} & 92.24 & 91.00 \\  \midrule
\begin{tabular}[c]{@{}l@{}}GPT2\\+Fine-Tuning\end{tabular}& 46.63 & 84.02 & 64.24 &  92.04 & 90.80\\  \midrule
\begin{tabular}[c]{@{}l@{}}GPT2\\+CoreLM\\+Fine-Tuning\end{tabular}& 48.11 & 84.16 & 64.48 & \textbf{92.40} & 90.88\\ \midrule

\begin{tabular}[c]{@{}l@{}}GPT2\\+CoreLM\\+Fine-Tuning\\+Coref\end{tabular}& \textbf{48.39} & 84.16 &  \textbf{64.56} & \textbf{92.40} & \textbf{91.97} \\ 
\bottomrule
\end{tabular}
\label{tab:acc_results}
\end{table*}

The Children's Book Test (CBT) \cite{hill2016cbt} was designed to evaluate LMs in different word categories. In comparison to GPT2, we report scores in all categories, i.e. Common Nouns (CN), Named Entities (NE), Verbs (V) and Prepositions (P). CBT is designed as a cloze test, in which a hidden word should be predicted, given ten possible options. We formulate this task as a language modeling task, similar to \citet{radford2019language}, in which we condition the sentence with each option and calculate its probability, choosing the one with the highest probability as the final prediction. Similarly to LAMBADA, we use a pretrained Coreference Resolution model to annotate the corpus with newly initialized entity mentions. 

\Cref{tab:acc_results} shows the results in terms of Accuracy with and without the use of coreference annotations. 
\paragraph{Without Coreference:}
Comparing the base model, with a fine-tuned version of the base model and CoreLM, it becomes obvious that the GUMBY does not generalize well with CBT. As a result, we note a drop in performance by just fine-tuning the model in all word categories. However, the CoreLM fine-tuned version results in better performance compared to the GPT2 fine-tuned model, with insignificant differences from the base model in almost all categories, with the exceptions of Verbs in which we notice a slightly better Accuracy. 

\paragraph{With Coreference:} 
Including coreference annotation to the CoreLM model results in small changes in performance in all categories. Specifically, there is no gain in performance when using entities in CN corpus variant. However, in the NE variant we achieve a 0.16\% increase. This very small increase is because in the majority of the cases, the cloze test answers are similar to the surface forms of the entities as found in the context and as such the correlation is very easy for model to make without the need of extra information. For the V variant of the corpus, there is no change to the performance while using coreference annotations as expected, while for the P variant, there is a increase of 1.09\% (statistically significant with t-test $p < 0.01$). This increase in prepositions is attributed to the accurate resolution of the nouns as mentions of entities that changed their representations. 

The performance changes, while using coreference annotation are indicative of the impact of entity representations in CoreLM. As our base model is capable of contextualizing each entry in CBT due to it's context window, we expect these improvements to be bigger in corpora which the information outside the context window of the model is required. 

\section{Error Analysis}
In our experiments, we showcased the effects of CoreLM on the base model, in different scenarios. In this section we investigate the cases in which the base model performed better than the CoreLM model and vice-versa. For that reason, we manually compared the predictions made between the two models, using the GUMBY Fine-Tuned GPT2 model as the deciding factor in our observations between the effects of CoreLM Fine-Tuning and simple Fine-Tuning. We limit our analysis on the CBT corpus which provides a meaningful way to evaluate the changes due to its word category variants.

In our analysis, we noticed that fine-tuning on GUMBY lead to wrong choices by the model in all categories, which were not made by the GPT2. However, when CoreLM was used with coreference annotations, 83\% of those cases were corrected. The vast majority of the corrected cases were in the Prepositions and Named Entities word categories, with only 19\% of the corrected ones in Verbs and Common Noun word categories. A positive correlation between corrected cases and correct coreference annotations was noticed, as cases in which CoreLM persisted on the wrong option also had different coreferent annotations than the correct sentence. 

In the cases where CoreLM performed better than both the fine-tuned and base model, we noticed the same correlation between coreference annotations that directly affected the available options. Unavailingly, a small number of cases in which, the nouns following a preposition or the Named Entities themselves were annotated in a wrong coreference cluster, lead to different probability distributions for the available options and eventually to making the wrong selection. Furthermore, comparing the predictions made by CoreLM with and without the use of coreference annotations, we notice that the majority of the errors persisted when the option was not directly annotated in a coreference cluster.

\section{Discussion}
Our approach takes advantage of PLMs and increases their performance by exploiting distant contextual information in the form of entity representations using coreference annotations. What is more, it can be used with and without the existence of such information, hence not hindering the flexibility of PLMs. Our experiements when Fine-Tuning GPT2 and the GPT2 with CoreLM on the same data, without using coreference information show that even when the fine-tuning data are not best suited for the downstream tasks, CoreLM maintains more of the original model's performance, making it a more resilient Fine-Tuning methodology. In addition, as CoreLM is very modular, it can be applied to the majority of LMs, including non-autoregressive approaches such as BERT.

In Language Modeling, our approach achieved significantly lower Perplexity in all corpora compared to the GPT2 Fine-Tuned version. What is more, GUMBY proved to be an ill-suited corpus to fine-tune for both LAMBADA and WikiText based on the post Fine-Tuning performance. Regardless, Fine-Tuning with CoreLM demonstrated significant gains, even compared to the pre Fine-Tuned model. 

In both LAMBADA and CBT, we show increase in Accuracy compared to GPT2 pre and post Fine-Tuning. Most notably, the Named Entity and Prepositions word types showed the biggest increase in CBT, with Common Nouns suffering in comparison compared to the pre Fine-Tuned version and Verbs attaining a slightly better Accuracy. Our error analysis highlights the effect of coreference annotations to these changes in performance. In all cases, Fine-Tuned GPT2 achieved lower scores in all word categories compared to CoreLM Fine-Tuned, which indicated that GUMBY is not a suitable fine-tuning corpus for these tasks. 

Using model-created coreference annotations during Zero-Shot evaluation did increase the performance slightly. While the performance increase is minor, the entity representations used were only initialized from the scope of each example as there was not long contextual information to take advantage off. What is more, coreference annotations increased the performance regardless of the information being within the context window in all the examples, indicating that further gains can be achieved by using coreference annotations extensively over large pieces of discourse. 

Unavailingly, our approach is based on the ability of other models to accurately predict coreference clusters so that CoreLM can exploit the coreferent mentions. The errors in the predicted clusters introduce noise, to both entity representations and the final model. What is more, maintaining a persistent set of entity representations, is computationally expensive and can be very burdening when considering a large collection of documents. As a result, an entity management mechanism, similar to the one used in \citet{toshniwal-etal-2020-learning}, would be required for CoreLM to scale efficiently to bigger document collections. 

\section{Related Work}
Early entity-aware LMs were trained from scratch with entity information available through the training process. Specifically, \citet{yang-etal-2017-reference} and \citet{ji-etal-2017-dynamic} both introduced models that used reference information with attention-based mechanisms to incorporate them into the model. \Citeposs{yang-etal-2017-reference} model made use of both intra-linguistic, coreferring mentions in text, and extra-linguistic, tables and lists, features through three different components, only creating learnable embeddings for intra-linguistic features. \Citeposs{ji-etal-2017-dynamic} model was focused only on intra-linguistic mentions, i.e. corefererring mentions, and introduced additional control variables indicating if the next token is part of an entity as well as the number of remaining entity tokens. Recently, \citet{kunz-hardmeier-2019-entity} extended \Citeposs{yang-etal-2017-reference} approach by using learnable entity embeddings. These approaches also have the ability to autoregressively predict the entity of the following word and constrain the word generation to a specific entity.

EnGen combines EntityNLM \citep{ji-etal-2017-dynamic} with S2SA \citep{sutskever2014sequence}, to train a generative language model that uses both previous sentence representations and entity representations in order to generate coherent text \citep{clark-etal-2018-neural}. In comparison to their approach, our approach handles entities and information flow differently. 
\citet{stylianou-vlahavas-2020-e} also introduced a Transformer-based approach towards incorporating learnable entity representations, using multi-head attentions inside all the Transformer blocks of the model. In comparison to past approaches, this approach was only focused on the effective use of entity information in a LM and did not predict the following entity information. However, all of these models required high quality annotated data to be trained from scratch which were limited and of specific genre \cite{kunz-hardmeier-2019-entity, stylianou-vlahavas-2020-e}.  

Other methods have focused on using only extra-linguistic information, ignoring pronouns and nominal mentions in text. ERNIE \cite{zhang-etal-2019-ernie} uses Knowledge Graphs (KG) to extract entities for identified named entities. However, ERNIE represents entity information using a pre-trained knowledge embedding model, trained on the used KG, and does not create dynamic entity representations.
\citet{liu-etal-2019-knowledge} use Knowledge Bases to learn word type embeddings based on the learned type representations. Similar to past approaches \citep{ji-etal-2017-dynamic,kunz-hardmeier-2019-entity} it can autoregressively constrain the prediction to a certain entity type. The type information is restricted using pre-defined vocabularies making the model less dynamic in its entity predictions. Both approaches have been found to be very effective for the tasks that were respectively designed, however they lack in expandability of their domain of application without requiring complete retraining.

\section{Conclusions and Future Work}
In this paper we presented CoreLM, a modular Fine-Tuning framework for PLMs to exploit model-created coreference annotations in order to create better mention representations and an overall better LM. In our experiments we showcased a performance increase when evaluating in a zero-shot setting, compared to the similarly fine-tuned model, even when the fine-tuning corpus did not generalize well to the end tasks. Our analysis shows that coreference annotations play a significant role in both Fine-Tuning and in downstream task performance, with correct annotations leading to better performance when used. 

In addition, our work helps in adding a new frontier to Coreference Resolution through the effective use of coreference annotations in Language Modeling. In this paper we showcased the effects of coreference annotation even when the information is within the context window of the model. Using coreference annotations can further lead to the decrease of the required context window and boost approaches like Shortformer \citep{press2020shortformer}, leading to better and more efficient LMs.

In the future we aim to create a more efficient approach to LM through the use of both intra-linguistic (Coreference) and extra-linguistic (KG) features. Undeniably, KGs provide a means for structured, high quality information that cannot be found in a single text. We believe that an information fusion from coreference annotation and graph nodes, along with short context window will not be computationally prohibitive and lead in better, information rich, LMs.

\section*{Acknowledgements}
This research is co-financed by Greece and the European Union (European Social Fund - ESF) through the Operational Programme “Human Resources Development, Education and Lifelong Learning” in the context of the project “Strengthening Human Resources Research Potential via Doctorate Research” (MIS-5000432), implemented by the State Scholarships Foundation (IKY).
\bibliography{anthology,custom}

\begin{thebibliography}{31}
\expandafter\ifx\csname natexlab\endcsname\relax\def\natexlab#1{#1}\fi

\bibitem[{Ba et~al.(2016)Ba, Kiros, and Hinton}]{ba2016layer}
Jimmy~Lei Ba, Jamie~Ryan Kiros, and Geoffrey~E Hinton. 2016.
\newblock Layer normalization.
\newblock \emph{arXiv preprint arXiv:1607.06450}.

\bibitem[{Beltagy et~al.(2020)Beltagy, Peters, and
  Cohan}]{Beltagy2020Longformer}
Iz~Beltagy, Matthew~E. Peters, and Arman Cohan. 2020.
\newblock Longformer: The long-document transformer.
\newblock \emph{arXiv:2004.05150}.

\bibitem[{Bengio et~al.(2003)Bengio, Ducharme, Vincent, and
  Janvin}]{bengio2003neural}
Yoshua Bengio, R{\'e}jean Ducharme, Pascal Vincent, and Christian Janvin. 2003.
\newblock A neural probabilistic language model.
\newblock \emph{The journal of machine learning research}, 3:1137--1155.

\bibitem[{Child et~al.(2019)Child, Gray, Radford, and
  Sutskever}]{child2019generating}
Rewon Child, Scott Gray, Alec Radford, and Ilya Sutskever. 2019.
\newblock Generating long sequences with sparse transformers.
\newblock \emph{arXiv preprint arXiv:1904.10509}.

\bibitem[{Clark et~al.(2018)Clark, Ji, and Smith}]{clark-etal-2018-neural}
Elizabeth Clark, Yangfeng Ji, and Noah~A. Smith. 2018.
\newblock \href {https://doi.org/10.18653/v1/N18-1204} {Neural text generation
  in stories using entity representations as context}.
\newblock In \emph{Proceedings of the 2018 Conference of the North {A}merican
  Chapter of the Association for Computational Linguistics: Human Language
  Technologies, Volume 1 (Long Papers)}, pages 2250--2260, New Orleans,
  Louisiana. Association for Computational Linguistics.

\bibitem[{Devlin et~al.(2019)Devlin, Chang, Lee, and
  Toutanova}]{devlin-etal-2019-bert}
Jacob Devlin, Ming-Wei Chang, Kenton Lee, and Kristina Toutanova. 2019.
\newblock \href {https://doi.org/10.18653/v1/N19-1423} {{BERT}: Pre-training of
  deep bidirectional transformers for language understanding}.
\newblock In \emph{Proceedings of the 2019 Conference of the North {A}merican
  Chapter of the Association for Computational Linguistics: Human Language
  Technologies, Volume 1 (Long and Short Papers)}, pages 4171--4186,
  Minneapolis, Minnesota. Association for Computational Linguistics.

\bibitem[{Gessler et~al.(2020)Gessler, Peng, Liu, Zhu, Behzad, and
  Zeldes}]{gessler2020gumby}
Luke Gessler, Siyao Peng, Yang Liu, Yilun Zhu, Shabnam Behzad, and Amir Zeldes.
  2020.
\newblock Gumby--a free, balanced, and rich english web corpus.
\newblock In \emph{Proceedings of The 12th Language Resources and Evaluation
  Conference}, pages 5267--5275.

\bibitem[{He et~al.(2016)He, Zhang, Ren, and Sun}]{he2016deep}
Kaiming He, Xiangyu Zhang, Shaoqing Ren, and Jian Sun. 2016.
\newblock Deep residual learning for image recognition.
\newblock In \emph{Proceedings of the IEEE conference on computer vision and
  pattern recognition}, pages 770--778.

\bibitem[{Hill et~al.(2016)Hill, Bordes, Chopra, and Weston}]{hill2016cbt}
Felix Hill, Antoine Bordes, Sumit Chopra, and Jason Weston. 2016.
\newblock \href {http://arxiv.org/abs/1511.02301} {The goldilocks principle:
  Reading children's books with explicit memory representations}.
\newblock In \emph{4th International Conference on Learning Representations,
  {ICLR} 2016, San Juan, Puerto Rico, May 2-4, 2016, Conference Track
  Proceedings}.

\bibitem[{Ji et~al.(2017)Ji, Tan, Martschat, Choi, and
  Smith}]{ji-etal-2017-dynamic}
Yangfeng Ji, Chenhao Tan, Sebastian Martschat, Yejin Choi, and Noah~A. Smith.
  2017.
\newblock \href {https://doi.org/10.18653/v1/D17-1195} {Dynamic entity
  representations in neural language models}.
\newblock In \emph{Proceedings of the 2017 Conference on Empirical Methods in
  Natural Language Processing}, pages 1830--1839, Copenhagen, Denmark.
  Association for Computational Linguistics.

\bibitem[{Kitaev et~al.(2020)Kitaev, Kaiser, and Levskaya}]{kitaev2020reformer}
Nikita Kitaev, {\L}ukasz Kaiser, and Anselm Levskaya. 2020.
\newblock Reformer: The efficient transformer.
\newblock \emph{arXiv preprint arXiv:2001.04451}.

\bibitem[{Kunz and Hardmeier(2019)}]{kunz-hardmeier-2019-entity}
Jenny Kunz and Christian Hardmeier. 2019.
\newblock \href {https://doi.org/10.18653/v1/W19-2803} {Entity decisions in
  neural language modelling: Approaches and problems}.
\newblock In \emph{Proceedings of the Second Workshop on Computational Models
  of Reference, Anaphora and Coreference}, pages 15--19, Minneapolis, USA.
  Association for Computational Linguistics.

\bibitem[{Liu et~al.(2019)Liu, Du, and Stoyanov}]{liu-etal-2019-knowledge}
Angli Liu, Jingfei Du, and Veselin Stoyanov. 2019.
\newblock \href {https://doi.org/10.18653/v1/N19-1117} {Knowledge-augmented
  language model and its application to unsupervised named-entity recognition}.
\newblock In \emph{Proceedings of the 2019 Conference of the North {A}merican
  Chapter of the Association for Computational Linguistics: Human Language
  Technologies, Volume 1 (Long and Short Papers)}, pages 1142--1150,
  Minneapolis, Minnesota. Association for Computational Linguistics.

\bibitem[{Liu et~al.(2020)Liu, Jiang, He, Chen, Liu, Gao, and Han}]{Liu2020On}
Liyuan Liu, Haoming Jiang, Pengcheng He, Weizhu Chen, Xiaodong Liu, Jianfeng
  Gao, and Jiawei Han. 2020.
\newblock \href {https://openreview.net/forum?id=rkgz2aEKDr} {On the variance
  of the adaptive learning rate and beyond}.
\newblock In \emph{International Conference on Learning Representations}.

\bibitem[{Merity et~al.(2016)Merity, Xiong, Bradbury, and
  Socher}]{merity2016pointer}
Stephen Merity, Caiming Xiong, James Bradbury, and Richard Socher. 2016.
\newblock Pointer sentinel mixture models.
\newblock \emph{arXiv preprint arXiv:1609.07843}.

\bibitem[{Paperno et~al.(2016)Paperno, Kruszewski, Lazaridou, Pham, Bernardi,
  Pezzelle, Baroni, Boleda, and Fern{\'a}ndez}]{paperno-etal-2016-lambada}
Denis Paperno, Germ{\'a}n Kruszewski, Angeliki Lazaridou, Ngoc~Quan Pham,
  Raffaella Bernardi, Sandro Pezzelle, Marco Baroni, Gemma Boleda, and Raquel
  Fern{\'a}ndez. 2016.
\newblock \href {https://doi.org/10.18653/v1/P16-1144} {The {LAMBADA} dataset:
  Word prediction requiring a broad discourse context}.
\newblock In \emph{Proceedings of the 54th Annual Meeting of the Association
  for Computational Linguistics (Volume 1: Long Papers)}, pages 1525--1534,
  Berlin, Germany. Association for Computational Linguistics.

\bibitem[{Press et~al.(2020)Press, Smith, and Lewis}]{press2020shortformer}
Ofir Press, Noah~A Smith, and Mike Lewis. 2020.
\newblock Shortformer: Better language modeling using shorter inputs.
\newblock \emph{arXiv preprint arXiv:2012.15832}.

\bibitem[{Radford et~al.(2019)Radford, Wu, Child, Luan, Amodei, and
  Sutskever}]{radford2019language}
Alec Radford, Jeffrey Wu, Rewon Child, David Luan, Dario Amodei, and Ilya
  Sutskever. 2019.
\newblock Language models are unsupervised multitask learners.
\newblock \emph{OpenAI blog}, 1(8):9.

\bibitem[{Rajbhandari et~al.(2020)Rajbhandari, Rasley, Ruwase, and
  He}]{Samyam2019zero}
Samyam Rajbhandari, Jeff Rasley, Olatunji Ruwase, and Yuxiong He. 2020.
\newblock Zero: Memory optimizations toward training trillion parameter models.
\newblock In \emph{Proceedings of the International Conference for High
  Performance Computing, Networking, Storage and Analysis}, SC '20. IEEE Press.

\bibitem[{Rasley et~al.(2020)Rasley, Rajbhandari, Ruwase, and
  He}]{rasley2020deepspeed}
Jeff Rasley, Samyam Rajbhandari, Olatunji Ruwase, and Yuxiong He. 2020.
\newblock \href {https://doi.org/10.1145/3394486.3406703} {Deepspeed: System
  optimizations enable training deep learning models with over 100 billion
  parameters}.
\newblock In \emph{Proceedings of the 26th ACM SIGKDD International Conference
  on Knowledge Discovery \& Data Mining}, KDD '20, page 3505–3506, New York,
  NY, USA. Association for Computing Machinery.

\bibitem[{Sennrich et~al.(2016)Sennrich, Haddow, and
  Birch}]{sennrich-etal-2016-neural}
Rico Sennrich, Barry Haddow, and Alexandra Birch. 2016.
\newblock \href {https://doi.org/10.18653/v1/P16-1162} {Neural machine
  translation of rare words with subword units}.
\newblock In \emph{Proceedings of the 54th Annual Meeting of the Association
  for Computational Linguistics (Volume 1: Long Papers)}, pages 1715--1725,
  Berlin, Germany. Association for Computational Linguistics.

\bibitem[{Strubell et~al.(2019)Strubell, Ganesh, and
  McCallum}]{strubell-etal-2019-energy}
Emma Strubell, Ananya Ganesh, and Andrew McCallum. 2019.
\newblock \href {https://doi.org/10.18653/v1/P19-1355} {Energy and policy
  considerations for deep learning in {NLP}}.
\newblock In \emph{Proceedings of the 57th Annual Meeting of the Association
  for Computational Linguistics}, pages 3645--3650, Florence, Italy.
  Association for Computational Linguistics.

\bibitem[{Stylianou and Vlahavas(2020)}]{stylianou-vlahavas-2020-e}
Nikolaos Stylianou and Ioannis Vlahavas. 2020.
\newblock \href {https://www.aclweb.org/anthology/2020.crac-1.1} {{E}.{T}.:
  Entity-transformers. coreference augmented neural language model for richer
  mention representations via entity-transformer blocks}.
\newblock In \emph{Proceedings of the Third Workshop on Computational Models of
  Reference, Anaphora and Coreference}, pages 1--10, Barcelona, Spain (online).
  Association for Computational Linguistics.

\bibitem[{Stylianou and Vlahavas(2021)}]{stylianou2021coref}
Nikolaos Stylianou and Ioannis Vlahavas. 2021.
\newblock \href {https://doi.org/https://doi.org/10.1016/j.eswa.2020.114466} {A
  neural entity coreference resolution review}.
\newblock \emph{Expert Systems with Applications}, 168:114466.

\bibitem[{Sutskever et~al.(2014)Sutskever, Vinyals, and
  Le}]{sutskever2014sequence}
Ilya Sutskever, Oriol Vinyals, and Quoc~V Le. 2014.
\newblock Sequence to sequence learning with neural networks.
\newblock In \emph{Advances in neural information processing systems}, pages
  3104--3112.

\bibitem[{Tay et~al.(2020)Tay, Dehghani, Bahri, and Metzler}]{tay2020efficient}
Yi~Tay, Mostafa Dehghani, Dara Bahri, and Donald Metzler. 2020.
\newblock Efficient transformers: A survey.
\newblock \emph{arXiv preprint arXiv:2009.06732}.

\bibitem[{Toshniwal et~al.(2020)Toshniwal, Wiseman, Ettinger, Livescu, and
  Gimpel}]{toshniwal-etal-2020-learning}
Shubham Toshniwal, Sam Wiseman, Allyson Ettinger, Karen Livescu, and Kevin
  Gimpel. 2020.
\newblock \href {https://doi.org/10.18653/v1/2020.emnlp-main.685} {Learning to
  {I}gnore: {L}ong {D}ocument {C}oreference with {B}ounded {M}emory {N}eural
  {N}etworks}.
\newblock In \emph{Proceedings of the 2020 Conference on Empirical Methods in
  Natural Language Processing (EMNLP)}, pages 8519--8526, Online. Association
  for Computational Linguistics.

\bibitem[{Vaswani et~al.(2017)Vaswani, Shazeer, Parmar, Uszkoreit, Jones,
  Gomez, Kaiser, and Polosukhin}]{vaswani2017transformers}
Ashish Vaswani, Noam Shazeer, Niki Parmar, Jakob Uszkoreit, Llion Jones,
  Aidan~N Gomez, \L~ukasz Kaiser, and Illia Polosukhin. 2017.
\newblock \href
  {https://proceedings.neurips.cc/paper/2017/file/3f5ee243547dee91fbd053c1c4a845aa-Paper.pdf}
  {Attention is all you need}.
\newblock In \emph{Advances in Neural Information Processing Systems},
  volume~30. Curran Associates, Inc.

\bibitem[{Yang et~al.(2019)Yang, Dai, Yang, Carbonell, Salakhutdinov, and
  Le}]{yang2019xlnet}
Zhilin Yang, Zihang Dai, Yiming Yang, Jaime Carbonell, Russ~R Salakhutdinov,
  and Quoc~V Le. 2019.
\newblock \href
  {https://proceedings.neurips.cc/paper/2019/file/dc6a7e655d7e5840e66733e9ee67cc69-Paper.pdf}
  {Xlnet: Generalized autoregressive pretraining for language understanding}.
\newblock In \emph{Advances in Neural Information Processing Systems},
  volume~32. Curran Associates, Inc.

\bibitem[{Yang et~al.(2017)Yang, Blunsom, Dyer, and
  Ling}]{yang-etal-2017-reference}
Zichao Yang, Phil Blunsom, Chris Dyer, and Wang Ling. 2017.
\newblock \href {https://doi.org/10.18653/v1/D17-1197} {Reference-aware
  language models}.
\newblock In \emph{Proceedings of the 2017 Conference on Empirical Methods in
  Natural Language Processing}, pages 1850--1859, Copenhagen, Denmark.
  Association for Computational Linguistics.

\bibitem[{Zhang et~al.(2019)Zhang, Han, Liu, Jiang, Sun, and
  Liu}]{zhang-etal-2019-ernie}
Zhengyan Zhang, Xu~Han, Zhiyuan Liu, Xin Jiang, Maosong Sun, and Qun Liu. 2019.
\newblock \href {https://doi.org/10.18653/v1/P19-1139} {{ERNIE}: Enhanced
  language representation with informative entities}.
\newblock In \emph{Proceedings of the 57th Annual Meeting of the Association
  for Computational Linguistics}, pages 1441--1451, Florence, Italy.
  Association for Computational Linguistics.

\end{thebibliography}
\bibliographystyle{acl_natbib}

\clearpage
\appendix

\begin{table*}[!ht]
\caption{Data example from the GUMBY corpus, as formatted for the task. Other corpora are similarly formatted, with multiple rows of entity annotation when coreference information were needed.}
\label{tab:gumby_example}
\resizebox{\textwidth}{!}{
\centering
\begin{tabular}{c|ccccccccccccccccccccc}
        \hline
        $X_{1:12}$ & The & prime & minister & of & Israel & , & Binyamin & Netanyahu & , & told & a & news \\
        $E_{1:12}$ & 11 & 11 & 11 & 11 & 11 & $\emptyset$ & 11 & 11 & $\emptyset$ & $\emptyset$ & 13 & 13 \\
        $E_{1:12}$ & $\emptyset$ & $\emptyset$ & $\emptyset$ & $\emptyset$ & 7 & $\emptyset$ & $\emptyset$ & $\emptyset$ & $\emptyset$ & $\emptyset$ & $\emptyset$ & $\emptyset$ \\
        \hline
    \end{tabular}
}
\end{table*}
\section{Experimental Setup}
\label{sec:experimental_setup}

In all our experiments we use the GPT2-small configuration with 124M parameters, with 12 layers and 12 attention heads each for our base model. We add one Entity-Gating layer after the base model's Transformer layers, which has a masked multi-headed entity attention layer with 12 heads and a 10\% dropout between layers. The gate flow rate ($\delta$) is set to 0.5. These hyperparameters were found to perform best from [6, 8, 10, 12, 16] number of heads and [0.2, 0.5, 0.7, 1] gate flow rate ($\delta$) after manual tuning.  

All datasets are tokenized using the pre-trained GPT2 tokenizer, which uses Byte Pair Encoding (BPE) \cite{sennrich-etal-2016-neural}. We also apply a simple de-tokenization based on author's responses in the official GitHub repository as the exact de-tokenizer used to achieve the published results has not been made available.\footnote{\href{https://github.com/openai/gpt-2/issues/131}{https://github.com/openai/gpt-2/issues/131}} We use the OpenAI LAMBADA split for evaluation and remove the The Jungle Book by Rudyard Kipling from CBT as it was found to be part of the GPT2 original training set \citep{radford2019language}. As such, all scores are based on our own experiments and in some cases vary (both positively and negatively) from the reported scores.

Our model has 132M parameters, a 6\% increase, after the addition of the Entity-Gating layer and the entity representations. It is fine-tuned on the GUMBY corpus for 10 epochs, with a batch size of 128. Rectified Adam \cite{Liu2020On} was used as the optimizer with 100 steps of warm up and a linearly decaying learning rate with a starting value of 1e-5. During Fine-Tuning, the entity representations are updated after every training step. We freeze all 12 GPT2 Transformer layers and apply gradients only to the input layers ($W_{e}$ and $W_{p}$), output layers (the language modeling head) and the Entity-Gating layer. During Zero-Shot evaluation we do not use any entity information and as such we discard the persistent entity representations. 

All experiments were run on a single Titan V 12GB graphics card, using  half precision floating-point format, Zero Stage 2 optimization \citep{Samyam2019zero} and DeepSpeed \citep{rasley2020deepspeed}. In this setup, fine-tuning takes approximately 8 hours, with no noticeable differences in terms of inference speed compared to GPT2. 

\section{Coreference Annotations}
\label{sec:coreference_annotations}
The vast majority of the datasets do not come with coreference annotations, a process which is very expensive and time consuming if it was to be done by human annotators. The same issue rises from the use of free text from web sources. In order to fully exploit our proposed framework, we uses the pre-trained Coreference Resolution model by \citet{toshniwal-etal-2020-learning} to create noisy coreference annotations for LAMBADA and Children's Book Test (CBT) corpora. 

Our approach does not have an entity-linking component with which the originally identified entities from the GUMBY corpus could be linked with newly identified entities in the other corpora. As such, the persistent entity representation used in the original GUMBY corpus were reset for each corpora. Hence, 
the resulting entity representation are not as descriptive as the GUMBY created ones as, there was little context involved and we only used the corpora for zero-shot evaluation, not allowing for iteratively creating richer entity representations.
\paragraph{LAMBADA:} 
No major preprocessing was required for the LAMBADA dataset. We did treat each entry in the dataset as a new document so that coreference annotations did not point to entities on other unrelated entities. The resulted entity representation were created as an average of the hidden representation of all the entity mention in that entry (\Cref{sec:entity_representations}), excluding the last word and its entity annotation which were to be predicted.
\paragraph{Children's Book Test:}
For CBT, we first formulated the corpus to fit our Language Modeling approach (\cref{sec:cbt}), conditioning each choice with one of 10 possible candidates and annotating the document as if the candidate was the answer in the cloze test. During evaluation, we predicted coreference clusters for the context conditioned with all possible candidate choices.
\end{document}